# GanitBench : A bi-lingual benchmark for evaluating mathematical reasoning in Vision Language Models

Ashutosh Bandooni
*School of Computer Science and Engineering*
*Vellore Institute of Technology*
Chennai, India
ashutosh.bandooni2021@vitstudent.ac.in

Brindha Subburaj
*School of Computer Science and Engineering*
*Vellore Institute of Technology*
Chennai, India
brindha.s@vit.ac.in

*Abstract*—Benchmarks for evaluating reasoning among Vision Language Models(VLMs) on several fields and domains are being curated more frequently over the last few years. However these are often monolingual, mostly available in English. Additionally there also is a lack of datasets available in Hindi on tasks apart from comprehension and translation. We introduce GanitBench, a tough benchmark consisting of 1527 vision-only questions covering several topics in Mathematics - available in languages English and Hindi. Collected from two major examinations from India, the JEE Advanced and the CBSE Boards examinations, this benchmark includes questions in the form of images comprising of figures essential to a question as well as text. We evaluate two closed source models for the same, in zero-shot Chain-of-Thought (CoT) and two-shot CoT settings. GPT-4o mini is found to be the more dominant model on the benchmark, with it's highest average accuracy being 38.15%. We also evaluate models through a "Double Lock" constraint, which brings down the performance of the models by considerable margins. We observe that two-shot CoT appears to be a more effective setting under this environment. Performance of the two VLMs also decreases when answering the same questions in the Hindi language. We hope to facilitate the inclusion of languages like Hindi in research through our work.

## I. Introduction

Since the inception of the Transformer architecture [1], Large Language Models have been a massive subject of interest in Deep Learning and Artificial Intelligence, excelling in tasks like Question-Answer reasoning, Machine Translation, Reading Comprehension etc. [2] even in a zero-shot manner. With the introduction of multimodal inputs in language models [3] [4], tasks requiring a combination of textual and visual contexts have started to emerge in more numbers [5]. Tasks for evaluating Vision Language Models combine Optical Character Recognition(OCR) with added inferential reasoning. Benchmarking is a vital part of research in any disciplinary field– helping researchers to evaluate their proposed works and methodologies and providing insights and pointers to look for in bringing about improvement to the same. Several Multimodal Language Models have been evaluated using wide variety of benchmarks available, with sources like [6] keeping track of their performances in a consolidated manner. Mathematical reasoning has been exclusively worked upon in the context of benchmarking, with several works evaluating Large Language Models on math problems ranging from basic elementary level questions all the way till complex university level questions. Most of these benchmarks appear to be in English, followed by Chinese being the second most prominent ones. We have observed that there are lower amount of datasets available for benchmarking related to mathematics in Hindi, which are not directly relying on translations through Neural Machine Translation or synthetic generation but rather on actual, untranslated sources – not only for Vision Language Models but also for Large Language Models. Having a mathematical benchmark with languages other than English can be beneficial for a more thorough evaluation of a model's reasoning capabilities. Our benchmark consists of mathematical questions from two prominent Indian examinations – JEE Advanced and the CBSE Class 10th and 12th Boards. Both tests are widely attempted by Indian students, and hence are archived and openly provided in both English and Hindi languages on the official websites of respective examinations [7] [8]. The questions are prompted in zero-shot CoT and two-shot CoT settings, for two proprietary and cost-effective models : GPT-4o mini and Claude 3 Haiku. The benchmark consists of questions having a large variety of answer types, including Multiple Choice questions(MCQs), numeric, list-based and many more with divisions within them on the basis of formatting of the ground-truth answers of the questions. The models are prompted with a system prompt, depending on the type of question, and are provided an image file which consists of the question in its entirety with all the information within the image itself, that is no extra information other than the basic system prompt is provided through texts. We conducted a "Double Lock" evaluation, which involved marking a specific question as "correctly solved" only if both the English and Hindi prompted answers came out correctly.

The primary contributions of this paper include:

 1. Providing a bi-lingual mathematical benchmark consisting of 1531 questions in English and Hindi, also consisting of diagrams and figures within them for visual contexts.

 2. Inferencing closed source VLMs in zero-shot and two-shot CoT settings, for evaluation on the benchmark

TABLE I: Comparison between Surveyed Works and GanitBench

| Sr. No | Name | Length of Dataset | Answer Type(s) | Multimodal / Vision Support? | Bi / Multilingual? |
|---|---|---|---|---|---|
| 1 | GSM8k | 8,500 | Numeric | No | No |
| 2 | MATH | 12,500 | Numeric | No | No |
| 3 | MathPrompter | 600 | Numeric | No | No |
| 4 | Geometry3k | 3,002 | Numeric | No | No |
| 5 | JEEBench | 515 | Numeric, Single Choice MCQ, Multiple Choice MCQ | No | No |
| 6 | NTSEBench | 2,728 | Single Choice MCQ | Yes | No |
| 7 | Mathify | 223 | Numeric | No | No |
| 8 | OlympiadBench | 8,476 | Numeric, Expression, Equation, Tuple, Interval, Theorem Proof | Yes | Yes |
| 9 | MathVista | 6,141 | Numeric, Single Choice MCQ | Yes | No |
| 10 | EXAMS-V | 20,932 | Single Choice MCQ | Yes | Yes |
| 11 | GanitBench | 1,527 | Numeric, Single Choice MCQ, Multiple Choice MCQ, List (Tuple), Numeric, Equation | Yes | Yes |

3. Deriving results under normal conditions and under "Double Lock" constraint

## II. RELATED WORKS

There are several works already existing involving benchmarking of LLMs and VLMs on a variety of domains. We summarise these works.

Mathematical reasoning on text : GSM8K [9], MATH [10], MathPrompter [11] and Geometry3k [12] all involve evaluating the mathematical reasoning capabilities of LLMs. GSM8K includes 8,500 grade-school math questions, complete with natural language solutions from expert annotators. The 6B and 175B variants of GPT-3 were used for evaluation, through both finetuning and verification training methods. MATH includes 12,500 questions collected from mathematical competitions like AMC 10 and AMC 12 across 7 subjects in mathematics, and is used on the GPT-2 and GPT-3 models. MathPrompter uses the MultiArith dataset in several combinations of zero-shot CoT and few-shot CoT settings on the GPT-3 175B model. Geometry3k includes 3002 geometry problems annotated in formal language, and was used on several neural network geometry solvers along with the proposed InterGPS based model. All these works lack visual contexts, are mono-lingual and majority of them include much easier questions in comparison to those provided in GanitBench.

Benchmarks with questions from Indian Examinations/Textbooks : JEEBench [13], NTSEBench [14] and Mathify [15] all utilise questions from examinations and textbooks made for students living in India by several Indian Educational authorities. JEEBench, a text-based benchmark, consists of 515 questions in Physics, Chemistry and Mathematics taken from the 2016-2023 editions of the Joint Entrance Examination(JEE) Advanced, evaluating open source LLMs and the proprietary GPT-4 model in several modes including zero-shot CoT, one-shot CoT, CoT self critique and CoT self consistency. NTSEBench, a multimodal benchmark, consists of 2,728 questions in Quantitative and Logical Aptitude from the National Talent Search Examination(NTSE), evaluating open source and proprietary LLMs and VLMs in zero-shot and few-shot settings. Mathify introduced a text-based dataset titled MathQuest consisting of 223 mathematical questions from the Class 11th and 12th NCERT Mathematics textbooks, on a open source models – with some of them specifically trained to handle mathematical reasoning. The work involved fine-tuning those models on several pre-existing mathematical datasets including MathQuest and evaluating their performance before and after fine-tuning.

The aforementioned benchmarks, although providing good results in their works, are completely mono-lingual. Additionally, NTSEBench only consists of Multiple Choice Questions, allowing a lack of subjective questions. JEEBench has questions with answers being numeric values, and MathQuest supports only numerals and tuples as answers. MathQuest and JEEBench also only contain questions in text, which voids several questions within textbooks and exam papers respectively that have figures to be included in the datasets.

Multimodal Math and Science benchmarks : OlympiadBench [16], MathVista [17], Exams-V [18] consist of questions involving both textual and visual contexts, across several regions of mathematics and science. OlympiadBench sources questions from several international Olympiads in maths and physics, along with the Chinese college entrance exam. The dataset includes 8,476 questions, is bi-lingual, and includes questions that would require figures and images to be known to the model for the entire context. The questions are open-ended, with some of them having a concrete final answer and the other being theorems to-be proved. The benchmark is used on open source and proprietary models, in a zero-shot setting. Theorem solving problems were manually verified for evaluation. MathVista comprises of 31 multimodal datasets, out of which 9 are entirely math-based, and is implemented in zero-shot CoT, two-shot CoT and two-shot Program of Thoughts (PoT) settings on 19 foundational models. Questions include both open-ended and multiple-choice answer formats. EXAMS-V involves 20,932 multimodal multiple choice questions in areas such as Social Science, Physics, Chemistry, Biology, Mathematics etc. in 11 different languages, collected from various examinations ranging from grade 4 to 12. VLMs and augmented models were used in zero-shot setting for evaluation.

EXAMS-V consists only multiple choice questions, leading to a less scrutinized evaluation. The dataset also doesn't include questions in Hindi, even after including JEE Advanced as a source. MathVista is entirely in English and hence does not provide insights about Large Multimodal Models' (LMMs) performance on the same questions in other languages.

Table I includes a comparison of GanitBench and works summarised in this section

### III. METHODOLOGY

#### A. Dataset Formation

*1) Data selection and extraction:* Using the official archive webpages for JEE Advanced and CBSE Boards Examinations, we were able to download PDFs of question papers available for each examination. For JEE Advanced, we downloaded papers for years ranging from 2016-2024 for English, and 20192024 for Hindi. The papers containing questions in Hindi were available as is, and were not annotated or translated through any tools. This ensures that the questions were written or transcribed authentically by native Hindi speakers. Questions from the Mathematics section were utilised for the dataset. For Class 10th Boards, we downloaded papers from the following math-related subjects : Math Basic and Math Standard. Math Basic is taken up by students who do not wish to pursue the subject of Mathematics for classes 11th and 12th, whereas Math Standard is to be compulsorily taken up by students who wish to do so. Math Basic, hence, contains questions of relatively easier difficulty when compared to Math Standard. For Class 12th Boards, we downloaded papers from the following math-related subjects : Mathematics and Applied Mathematics. Applied Mathematics focuses more on usage of mathematics in real world applications such as in engineering, physics, medicine etc. whereas Mathematics focuses more on principles and theories. As an example, the following is a question from the dataset, dealing with the concept of Compound Interest in Applied Mathematics :

*Amrita buys a car for which she makes a down payment of* 2,50,000 *and the balance is to be paid in 2 years by monthly instalments of* 25,448 *each. If the financer charges interest at the rate of 20% p.a., find the actual price of the car.*

From Class 10th, we used 5 papers for Math Basic and 6 papers for Math Standard across 3 years. From Class 12th, we used 3 papers for Applied Mathematics and 5 papers for Mathematics across the same year range. Questions were manually extracted through the Windows in-built Snip tool, and saved to their appropriate folders. Each image was named using a specific naming convention, which allowed for regularity while parsing folders through automation scripts to be used in later steps. Some image files for questions in Hindi in both JEE and Boards questions contained words/statements in English, specifically to aid students taking up the exam in Hindi to recognise the meaning of specific uncommon words. We manually censor such words using Paint tool in Windows, to ensure the models receive almost the entirety of the question in Hindi. Figure 1 provides an example for the same.

(a) A question available in Hindi

माना कि $\omega \neq 1$ एकक का एक घनमूल ▬▬▬ है| तब समुच्चय ▬▬▬

$$\{|a + b\omega + c\omega^2|^2 : a, b, c \text{ भिन्न अशून्य पूर्णांक} \text{▬▬▬} \text{ हैं}\}$$

का निम्नतम ▬▬▬ बराबर _____

(b) Same question after censoring non-essential English words

Fig. 1: A question in Hindi from JEE advanced a) in its original form and b) in its final and dataset ready form

*2) Annotating answers and metadata:* We created CSV files in two different formats, one for CBSE and one for JEE and wrote scripts for filling the CSV files up with metadata related to the corresponding examination about the images appropriately. The previously mentioned naming convention allowed for most metadata to be filled in rows automatically, since each file's name contained all necessary information for the same. The columns common among both formats of CSVs include the name of the image file, the correct final answer for the question, the type of question or the format required for the answer, the year of the question paper and language. Using verified sources available on the internet, we manually fill in the final answers for the questions. We maintain standardised formats for each question type, determined on the basis of the answer type asked in the question itself and the way it has been represented in the solutions. We discuss more about the eligible answer formats in the Results sections.

#### B. Experimentation

*1) Setup:* All scripts / code snippets pertaining to the work were written and executed using local Juypter Notebooks, on a laptop containing 16 GB of RAM and a 1 TB SSD storage ROM. Annotations and corrections in annotations to CSV files were done with the help of Microsoft Excel.

माना कि $\omega \neq 1$ एकक का एक घनमूल (a cube root of unity) है| तब समुच्चय (set)

$$\{|a + b\omega + c\omega^2|^2 : a, b, c \text{ भिन्न अशून्य पूर्णांक (distinct non-zero integers) हैं}\}$$

का निम्नतम (minimum) बराबर _____

*2) Prompting Strategies Used:* The work utilises two known strategies for prompting VLMs : Zero-Shot Chain-ofThought(CoT) and Two-Shot CoT. Chain-of-Thought generally involves giving a model specific instructions to make it generate reasoning chains one by one. These instructions can be a part of the system prompt itself, and works best in a hybridised manner – paired either with zero-shot or few-shot techniques.

Similiar to [19], we created system prompts that instructed the model to provide its answers in a sequential manner, and emphasised on the usage of theorems and methods for reaching the final answer. An example of system prompts used in our work is given in Figure 2. It provides the Hindi and English prompt used for questions having their final answers in the form of a list of values containing symbolic expressions. We use specific tags within the prompt while specifying the answer format just for making it distinguishable for the model. In zero-shot setting, the system prompt associated with the question along with the image file itself are both simply inputted as one unit. In two-shot setting, along with the system prompt and the image file, two example sets of questions along with their annotated and verified solutions are inputted as a stream of previous messages – intended to make the model learn from those examples as it's own expected output for those questions. We created two examples for each question type for both JEE Advanced and Boards questions, and stored them in JSON files to be used while inferencing.

*3) Inferencing and extraction of answers:* We utilise two closed-source proprietary models, Claude 3 Haiku and GPT4o mini for evaluation. Both models have significantly cheaper price-per-token rates compared to their heftier counterparts within their model families. We set the maximum response length to 2500 and temperature to 0.5 for both the models. We use API's provided for both models for inferencing. Images are encoded to base64 format and then used as inputs in the specific JSON format required by the API to input multimodally, as a list of messages under the messages parameter. We run scripts to parse CSV files, retrieve the image name from the respective column and locate the image in the respective folder. After which, it is converted to a base64 encoded string and then inferenced. The system prompt is determined by the value in the column related to the answer type, with mappings defined for the same in the script. For two-shot setting, we additionally also input a dictionary consisting of image(question)-explanation pairs to be inferenced to the models. The responses are then saved in their respective CSV

Fig. 2: System prompts used for questions with List(Latex) answer format, given in English and Hindi

file.

After finishing the process of inferencing entirely, we manually extracted the answers from the responses. This was done due to irregular patterns found in final answers, which could not have been automated through regex patterns as well as problems related to parsing of symbolic expressions. We made sure to extract only those answers which followed the particular instructions and formats required for their answer type. For those not following the formats, we simply wrote "N.A" or "None" against their column in the record. For example, if a question is supposed to have a list of values consisting of symbolic expressions in latex (fractions, logarithms etc.), if the extracted answer seen does not contain the answers in latex, we considered it incorrect. We gave some leeway for non-simplified expressions, if it was possible to reduce them within 2-3 steps.

*4) Scoring and Evaluation:* For JEE Questions, we employed an automatic scoring system that gave a score of either 0 or 1 for each question. If the extracted answer was entirely correct with respect to the rules for that question type's evaluation, score granted to the question was 1, otherwise 0. We do not consider partial scoring in our work, and instead go for a stricter criteria. For Boards questions, we manually score by comparing extracted and correct answer. Accuracy can be represented by :

$$\text{Accuracy} = \frac{\sum_{i=1}^{N} \text{Score}_i}{N} \qquad (1)$$

Where N is the number of questions in the portion of the dataset being used. We find out accuracies of models firstly individually, that is in "normal" mode for each language and examination, and then in a "Double Lock" mode. Under this mode, a specific question is considered to have been correctly answered if and only if scores for both the English and Hindi versions were found to be 1, indicating a correct solution. This allows us to examine the effect of languages on reasoning, and provides a harsher environment for evaluation

TABLE II: Exam-Language split for the dataset

| Division/Language | Hindi | English | Total |
|---|---|---|---|
| Boards | 518 | 518 | 1036 |
| JEE | 198 | 293 | 491 |
| Total | 718 | 813 | 1527 |

TABLE III: Question type wise split for JEE Advanced questions

| Type/Language | English | Hindi | Total |
|---|---|---|---|
| Single choice | 68 | 38 | 106 |
| Multiple choice | 103 | 64 | 167 |
| Numerical | 68 | 52 | 120 |
| Integer | 54 | 44 | 98 |
| Total | 293 | 198 | 491 |

TABLE IV: Question type wise split for Boards Questions

| Type/Language | English | Hindi | Total |
|---|---|---|---|
| List(Latex) | 17 | 17 | 34 |
| Numerical | 114 | 114 | 228 |
| Numerical(Latex) | 56 | 56 | 112 |
| List | 35 | 35 | 70 |
| Assertion Reasoning | 28 | 28 | 56 |
| List (Coordinates) | 12 | 12 | 24 |
| Single Choice MCQ | 251 | 251 | 502 |
| Equation | 5 | 5 | 10 |
| Total | 518 | 518 | 1036 |

IV. RESULTS

A. Dataset Details

The final dataset amasses 1527 images, including questions in both Hindi and English. An exam-wise split of the questions has been provided in Table II. There are four answer types in JEE Advanced:

- Single Choice MCQ – Only one out of four options is correct. An example of a formatted answer would be "A".
- Multiple Choice MCQ – One or more options out of four is/are correct. An example of a formatted answer would be "[A,B,C]".
- Integer – The answer is an integer value, ranging from $-\infty$ to $+\infty$. An example of a formatted answer would be "5".
- Numeric – The answer is a decimal value, supposed to be rounded off to the second place. An example of a formatted answer would be "5.67".

For Boards, we defined eight different answer types:

1) Single Choice MCQ – Only one out of four options is correct. An example of a formatted answer would be "A".
2) Assertion Reasoning – Given an assertion and a reasoning statement, find out the relationship between the two, indicated through a single-choice MCQ. An example of a formatted answer would be "A".
3) Numerical – The answer is a simplified numeric (including decimals) value, ranging from –inf to +inf. An example of a formatted answer would be "5712" or "4.567".
4) Numerical(Latex) – The answer is a numeric value, either in the form of an irrational or a fraction, used to represent symbolic expressions. An example of a formatted answer would be "$\frac{4}{5}$" or "$2000\sqrt{3}$".
5) List – The answer is a list of simplified numeric values, preferably in the order of their appearance in the question. An example of a formatted answer would be "[12,34,156]".
6) List(Coordinates) – The answer is a list of coordinates/variables, either in 2 or 3 dimensions or as a variable linear equation. An example of a formatted answer would be "[0,-1]" or "[3,4,5]". These do not include symbolic expressions.
7) List(Latex) – The answer is a list of values consisting of symbolic expressions. An example of a formatted answer would be "$[\frac{1}{5}, \frac{2}{5}]$".
8) Equation - The answer is an equation. An example of a formatted answer would be $y = x^2 - \log x + C$

A format-wise split of questions has been provided in Table III and Table IV

TABLE V: Results of Normal Mode

|  | JEE | | BOARDS | | AVERAGE. |
|---|---|---|---|---|---|
|  | EN | HI | EN | HI | |
| GPT-4o mini (Zero-shot CoT) | 0.2111 | 0.1010 | 0.7277 | 0.4864 | 0.3815 |
| GPT-4o mini (Two-shot CoT) | 0.2013 | 0.1060 | 0.7239 | 0.4710 | 0.3755 |
| Claude 3 Haiku (Zero-shot CoT) | 0.1262 | 0.1414 | 0.3571 | 0.1563 | 0.1952 |
| Claude 3 Haiku (Two-shot CoT) | 0.0989 | 0.0959 | 0.3223 | 0.1891 | 0.1765 |

B. Performance of Models

Table V provides results obtained in normal mode. GPT4o mini appears to be performing better across both JEE and Boards questions over Claude 3 Haiku, with the model scoring an average of 38.15% in the zero-shot CoT setting compared to 19.52% from the latter. Two-shot CoT has not necessarily led to an increase in performance, with accuracies under the setting usually decreasing for both models relatively to that under zero-shot CoT. Performance of the models on Hindi questions decreased in comparison to their English equivalents, with the zero-shot CoT inferenced Claude 3 Haiku being the only exception in case of questions from JEE, seeing an increase in accuracy from 12.62% to 14.14%.

TABLE VI: Results after applying Double Lock constraint

| Model | JEE | BOARDS | AVERAGE |
|---|---|---|---|
| GPT-4o mini (Zero-shot CoT) | 0.0000 | 0.4420 | 0.2210 |
| GPT-4o mini (Two-shot CoT) | 0.0555 | 0.4111 | 0.2333 |
| Claude 3 Haiku (Zero-shot CoT) | 0.0353 | 0.0849 | 0.0601 |
| Claude 3 Haiku (Two-shot CoT) | 0.0202 | 0.1061 | 0.0632 |

When applying the Double Lock constraint, accuracies for models took further hits and are observed to be lesser - with the highest accuracy decreasing from 38.15% to 23.33%. On average, two-shot CoT setting provided higher accuracies from both the models than in zero-shot CoT setting. We observe that models become less accurate when the scoring criteria considers one question having the correct generated solution in more than one language, inferring that VLMs do struggle with reasoning when dealing with languages other than English. Accuracies of the models after applying the Double Lock constraint are provided in Table VI

## V. Conclusion

We present a bi-lingual benchmark consisting of mathematical questions across several topics such as Calculus, Coordinate Geometry, Probablity and Statistics, Trigonometry, Inferential Statistics and many more. We provide observations and results about the performance of two VLMs on the benchmark. Under normal conditions, GPT 4o-mini scores the highest overall accuracy of 38.15% and 37.55% in zero-shot and two-shot CoT settings respectively, with the accuracies of the Claude 3 Haiku model being almost half of those. The models answer questions from the Boards examination more accurately than from JEE Advanced. When applying the "Double Lock" constraint for finding out accuracies, results indicated a steep drop in both models' performance – with one instance containing a nil percent accuracy figure. The constraint falls short in deciding whether erroneous responses of VLMs are caused by lack of linguistic understanding or by flawed mathematical judgement - rather helping in explaining how well models understand versions of the same question in different languages. The proposed benchmark can be expanded to include more subjects such as Physics and Chemistry from the currently used examinations, topics from other prominent examinations and spoken languages used within India. We hope to see more benchmarks and datasets consisting of questions in Hindi in upcoming works, and more works involving mathematical reasoning in non-English languages.